%% file: variational_koopman.tex
\documentclass{article}
\usepackage{ijcai19}

\usepackage{times}
\usepackage{soul}
\usepackage{url}
\usepackage[hidelinks]{hyperref}
\usepackage[utf8]{inputenc}
\usepackage[small]{caption}
\usepackage[utf8]{inputenc} 
\usepackage[T1]{fontenc}    
\usepackage{hyperref}       
\usepackage{url}            
\usepackage{booktabs}       
\usepackage{amsfonts}       
\usepackage{nicefrac}       
\usepackage{microtype}      
\usepackage{graphics}
\usepackage{pgfplots}
\usepackage{pgfplotstable}
\usepgfplotslibrary{fillbetween}
\usepackage{tikz}
\usepackage{bm}
\usepackage{xfrac}
\usepackage{float}
\usepackage[separate-uncertainty=true,detect-mode=true]{siunitx}    
\usepackage{color}
\usepackage{subcaption}
\usepackage{algorithm}
\usepackage[noend]{algpseudocode}
\usepackage{mathtools}
\usepackage{amssymb}
\usepackage{amsmath}
\usepackage{standalone}
\usepackage[keeplastbox]{flushend}
\usepackage[capitalize]{cleveref}
\usepackage{graphicx}
\usepackage{multirow}

\usetikzlibrary{arrows}
\usetikzlibrary{arrows.meta}
\usetikzlibrary{bayesnet}
\usetikzlibrary{fit}
\usetikzlibrary{matrix}
\usetikzlibrary{positioning}
\usetikzlibrary{intersections}
\usetikzlibrary{graphs}
\usetikzlibrary{calc}
\usetikzlibrary{patterns}
\usetikzlibrary{decorations.pathreplacing}
\usepgfplotslibrary{groupplots}

\pgfplotsset{/pgfplots/ybar legend/.style={
    /pgfplots/legend image code/.code={%
       \draw[##1,/tikz/.cd,yshift=-0.25em]
        (0cm,0cm) rectangle (3pt,0.8em);},
   },
   ylabsh/.style={ 
     axis y label/.style={
      at={(0,0.5)}, xshift=#1, rotate=90
      }
  },
}

\pgfplotsset{compat=newest}

\pgfplotsset{every axis legend/.append style={legend cell align=left}}

\definecolor{colorBluish}{RGB}{101,161,216}
\definecolor{colorBluePastel}{RGB}{240,240,255}
\definecolor{colorGreenish}{RGB}{0,130,0}
\definecolor{colorGreenPastel}{RGB}{240,250,240}
\definecolor{colorRedish}{RGB}{140,21,21}
\definecolor{colorRedPastel}{RGB}{255,240,240}
\definecolor{colorOrangish}{RGB}{237,125,49}
\definecolor{colorOrangePastel}{RGB}{213,169,143}
\definecolor{colorPurplePastel}{RGB}{216,191,216}

\newcommand{\T}{{^\intercal}}

\newcommand{\R}{\mathbb{R}}
\DeclareMathOperator*{\E}{\mathbb{E}}

\newcommand{\x}{\mathbf{x}}
\newcommand{\us}{\mathbf{u}}
\newcommand{\g}{\mathbf{g}}
\newcommand{\z}{\mathbf{z}}

\title{Deep Variational Koopman Models: Inferring Koopman Observations\\ for Uncertainty-Aware Dynamics Modeling and Control}

\author{
Jeremy Morton$^1$\and
Freddie D. Witherden$^2$\And
Mykel J. Kochenderfer$^1$\\
\affiliations
$^1$Department of Aeronautics and Astronautics, Stanford University\\
$^2$Department of Ocean Engineering, Texas A\&M University\\
\emails
jmorton2@stanford.edu, fdw@tamu.edu, mykel@stanford.edu
}

\begin{document}

\maketitle

\begin{abstract}
Koopman theory asserts that a nonlinear dynamical system can be mapped to a linear system, where the Koopman operator advances observations of the state forward in time.
However, the observable functions that map states to observations are generally unknown.
We introduce the Deep Variational Koopman (DVK) model, a method for inferring distributions over observations that can be propagated linearly in time. 
By sampling from the inferred distributions, we obtain a distribution over dynamical models, which in turn provides a distribution over possible outcomes as a modeled system advances in time.
Experiments show that the DVK model is effective at long-term prediction for a variety of dynamical systems.
Furthermore, we describe how to incorporate the learned models into a control framework, and demonstrate that accounting for the uncertainty present in the distribution over dynamical models enables more effective control.

\end{abstract}

\section{Introduction}\vspace{-0.25em}

In order to analyze, control, and predict the evolution of dynamical systems, we require knowledge about their governing equations. 
For many complex systems of interest, the exact governing equations are either unknown or are prohibitively expensive to accurately evaluate.
These challenges have inspired recent interest in learning system dynamics directly from data.
In particular, neural network dynamics models have garnered widespread attention due to their ability to model complex functions with high-dimensional inputs such as image data~\cite{krishnan2017structured,rangapuram2018deep,moerland2017learning,karl2017deep,fraccaro2017disentangled}.

Data-driven dynamics modeling is of particular interest to the field of reinforcement learning (RL), where the goal is to automatically learn control policies that satisfy predefined objectives.
Model-based RL algorithms, which attempt to explicitly model environment dynamics, have the potential to solve complex tasks while requiring significantly less experience than model-free algorithms. 
However, the difficulty of constructing accurate data-driven dynamics models has so far allowed model-free approaches to outperform model-based approaches on many problems.
Nonetheless, recent work has demonstrated that planning algorithms combined with neural network dynamics models can achieve strong performance on a variety of tasks while requiring less environmental interaction than state-of-the-art model-free RL algorithms~\cite{chua2018deep,hafner2018learning}.

The exact form of a dynamics model has strong implications for how easily the model can be incorporated into a control or planning framework.
Neural networks are nonlinear functions, meaning neural dynamics models might not be well suited for many control methods designed for linear systems.
For this reason, approaches such as E2C~\cite{watter2015embed} and RCE~\cite{banijamali2018robust} train neural networks to map states to a latent space where the dynamics can be evolved according to locally linear models that enable action selection through iLQR.
Koopman theory~\cite{koopman1931hamiltonian} offers an alternative viewpoint through which nonlinear dynamics can be mapped to linear dynamics.
It posits the existence of a linear operator that acts on observable functions of the state to advance them forward in time.
The exact form of the observable functions is usually not known, but recent work has sought to learn them automatically using neural networks~\cite{lusch2018deep,takeishi2017learning,li2017extended,otto2017linearly}.
Furthermore, it has been shown that data-driven models that leverage Koopman theory can be used for control in a wide array of applications~\cite{kaiser2017data,korda2018linear,morton2018deep}.

In this work, we introduce the Deep Variational Koopman (DVK) model, a method for inferring distributions over Koopman observations that can be propagated linearly in time.
Our method requires the training of a single neural network model, but enables the sampling of an ensemble of linear dynamics models in the space of observations.
Taken together, this model ensemble effectively provides a distribution over the system dynamics.
In evaluations on benchmark problems, we demonstrate that DVK models can be used for accurate long-term prediction with reasonable uncertainty estimates.
Additionally, we explain how the linear model ensembles can be easily incorporated into an existing control framework, and we empirically demonstrate that controller effectiveness improves as the size of the model ensemble grows.\vspace{-0.5em}

\section{Dynamics Modeling}\vspace{-0.25em}

In this section, we provide background on the Koopman operator and its relation to forced dynamical systems.
Subsequently, we derive an objective function for inferring distributions over Koopman observations, which yields a practical training procedure for inferring Koopman observations from data.
\subsection{The Koopman Operator}
Consider a nonlinear discrete-time dynamical system subject to control inputs described by $\mathbf{x}_{t+1} = F(\mathbf{x}_t, \mathbf{u}_t)$, where $\mathbf{x}_t \in \mathbb R^n$ and $\mathbf{u}_t \in \R ^p$.
Koopman theory asserts that there exists an infinite-dimensional linear operator $\mathcal K$ that advances all observable functions $h$ of the state and control inputs forward in time. Under the assumption that the control inputs are not evolving dynamically, this update equation takes the form~\cite{proctor2018generalizing}:
\begin{equation}
\mathcal K h(\mathbf{x}_t, \mathbf{u}_t) = h(F(\mathbf{x}_t, \mathbf{u}_t), \bm{0}) = h(\mathbf{x}_{t+1}, \bm{0}).
\end{equation}
If there exist a finite number of observable functions $\{h_1, \ldots, h_m\}$ that span a subspace $\mathcal H$ such that $\mathcal K h \in \mathcal H$ for all $h \in \mathcal H$, then $\mathcal H$ is considered to be an invariant subspace and $\mathcal K$ becomes a finite-dimensional operator $K$.

In this work, we assume that the observables take the form $h(\mathbf{x}_t, \mathbf{u}_t) = g(\mathbf{x}_t) + L \mathbf{u}_t$, where $g(\mathbf{x}_t)$ represents an observation of state $\mathbf{x}_t$ and $L \in \R^{1 \times p}$ is a matrix.  
Under the assumption of a finite-dimensional Koopman operator and defining the vector-valued observables $\mathbf{h} = [ h_1, \ldots, h_m]\T$ and $\mathbf{g} = [ g_1, \ldots, g_m]\T$, we have the update equation:
\begin{equation}
	\begin{split}
		\g(\x_{t+1}) & = K \mathbf{h}(\x_t, \us_t) = \begin{bmatrix}
		A &  B
	\end{bmatrix} \begin{bmatrix}
		\g(\x_t)\\
		\us_t
	\end{bmatrix}\\	
	& = A \g(\x_t) + B \us_t.
	\end{split}
    \label{eq:forward_time}
\end{equation}
The above expression describes the forward-time evolution of the observations $\g(\x_t)$; if $A$ is invertible we can likewise describe the reverse-time evolution as
\begin{equation}
\g(\x_t) = A^{-1}\left(\g(\x_{t+1}) - B \us_t\right).
\label{eq:reverse_time}
\end{equation}

Consider a sequence of control inputs $\mathbf{u}_{1:T-1}$ applied to a system with a finite-dimensional Koopman operator, resulting in a sequence of states $\mathbf{x}_{1:T}$.
Define the matrices $Z \in \R^{(m+p) \times (T-1)}$, $Y \in \R^{m \times (T-1)}$ as:
\begin{equation}
\begin{split}
    Z &= \begin{bmatrix}
        \g(\x_1) & \g(\x_2) & \ldots & \g(\x_{T-1})\\
        \us_1 & \us_2 & \ldots & \us_{T-1}
\end{bmatrix} \\ 
    Y &= \begin{bmatrix}
        \g(\x_2) &  \g(\x_3) & \ldots & \g(\x_T)
\end{bmatrix}.
\end{split}
\end{equation}
Under the assumptions outlined above, we can recover $A$ and $B$ through $[A \; B] = YZ^\dagger$ for sufficiently long sequences, where $Z^\dagger$ is the Moore-Penrose pseudoinverse of $Z$.
Unfortunately, the true form of the observables is generally unknown.
In this work, we attempt to infer the sequence of observations $\mathbf{g}(\mathbf{x}_1), \ldots, \mathbf{g}(\mathbf{x}_T)$ based on a sequence of control inputs and the time evolution of the state of a dynamical system.
Note that we do not model the functions $g$ directly, but instead attempt to infer the value of the observations.

\subsection{Inference Procedure}

\begin{figure}[t]
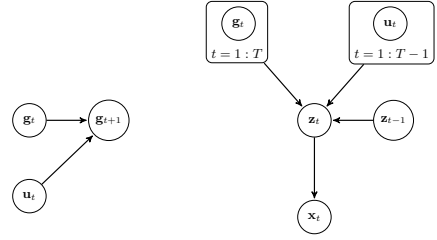

	\begin{center}
		\includestandalone[width=0.66\linewidth]{graph}
		\caption{Graphical model. Each observation $\g_t$ is a function of the previous observation and control input. Because the dynamical model that governs the time evolution of $\z_{1:T-1}$ is derived directly from $\g_{1:T}$ and $\us_{1:T-1}$, influence flows from these variables to $\z_{1:T-1}$.}\vspace{-1.5em}
		\label{fig:graph}
	\end{center}
\end{figure}

Consider a system subjected to a sequence of control inputs $\us_{1:T-1}$, causing it to traverse a set of states $\x_{1:T}$.
We assume there exist observations of the states $\g(\x_t)$ such that the system can be simulated linearly in time as outlined in \cref{eq:forward_time} and \cref{eq:reverse_time}.
Furthermore, we assume that the sequence is sufficiently long such that, when we form the matrices $Z$ and $Y$ as defined above, we have $[A \;  B] = YZ^\dagger$, i.e. we can find the true $A$ and $B$ matrices directly from the observations and control inputs.
Let $\g_t$ be a latent variable representing the observation $\g(\x_t)$.
Additionally, we introduce $\z_{1:T}$ as a set of latent variables that enforce that multi-step predictions made with the derived dynamics model allow for accurate reconstructions of the states $\x_{1:T}$.
Because we would expect prediction error to grow with time, we simulate the $\z_t$'s backward in time such that the lowest reconstruction error is generally obtained at time $T$, which in turn will allow more accurate predictions for how a system will evolve in future, unobserved time steps.
The values of $\z_{1:T}$ can be determined through:
\begin{equation}
\z_T = \g_T, \;\; \z_{t} = A^{-1}\left(\z_{t+1} - B \us_t \right).
\label{eq:zs}
\end{equation}
Finally, we assume that $\x_t = \g^{-1}(\z_t)$, i.e. the states $\mathbf{x}_t$ are generated by inverting the observable function $\mathbf{g}(\cdot)$.
\Cref{fig:graph} shows the graphical model for this problem.
	
We desire a model that maximizes the likelihood assigned to a sequence of observed states $\x_{1:T}$ conditioned on actions $\us_{1:T-1}$.
In modeling this density, we need to account for the presence of latent variables $\g_{1:T}$ and $\z_{1:T}$.
We can write the expression for the likelihood $L = p(\x_{1:T} \mid \us_{1:T-1})$ in terms of the observed and latent variables as:
\begin{equation}
L = \int p(\g_{1:T}, \z_{1:T}, \x_{1:T} \mid \us_{1:T-1}) d \g_{1:T} d \z_{1:T}.
\end{equation}
By the chain rule, the integrand can be factored into:
\begin{equation}
\begin{split}
p &(\g_{1:T}, \z_{1:T}, \x_{1:T} \mid \us_{1:T-1}) = p(\g_{1:T} \mid \us_{1:T-1})\, \times\\
&p(\z_{1:T} \mid \g_{1:T}, \us_{1:T-1})\, p(\x_{1:T} \mid \g_{1:T}, \z_{1:T}, \us_{1:T-1})
\end{split}
\end{equation}
Each term can be simplified using the conditional independence assumptions encoded by the graphical model.
The first term can be simplified to:
\begin{equation}
    p(\g_{1:T} \mid \us_{1:T-1}) = p(\g_1) \prod\nolimits_{t=2}^T p(\g_t \mid \g_{t-1}, \us_{t-1}).
\end{equation}
Each factor in the above expression can be thought of as a (conditional) prior over an observation at time $t$.
The second term in the integrand describes the distribution over variables $\z_{1:T}$, whose values can be determined exactly using \cref{eq:zs} if $\g_{1:T}$ and $\us_{1:T-1}$ are known. 
Thus, we find:
\begin{equation}
\begin{split}
p(\z_{1:T} \mid \g_{1:T}, & \us_{1:T-1})  = \delta(\z_T \mid \g_T) \, \times \\
& \prod\nolimits_{t=1}^{T-1}\delta(\z_t \mid \z_{t+1}, \g_{1:T}, \us_{1:T-1}),
\end{split}
\end{equation}
where $\delta(\cdot \mid \cdot)$ represents a deterministic relationship. 
From the structure of the graphical model, the last term becomes:
\begin{equation}
p(\x_{1:T} \mid \g_{1:T}, \z_{1:T}, \us_{1:T-1}) =  \prod\nolimits_{t=1}^T p(\x_t \mid \z_t).
\end{equation}

Even with these simplifications, evaluating the likelihood expression is generally intractable because it requires marginalizing over the latent variables.
Therefore, instead of optimizing this objective directly, we use variational inference to optimize a lower bound.
We introduce $q(\g_{1:T}, \z_{1:T} \mid \x_{1:T}, \us_{1:T-1})$, an approximation of the true posterior distribution over the latent variables. Multiplying and dividing the likelihood expression by this quantity, taking the logarithm of both sides, and invoking Jensen's inequality, we find a lower bound on the log-likelihood $\ell = \log p(\x_{1:T} \mid \us_{1:T-1})$:
\begin{equation}
\begin{split}
	\ell \geq & \E \left[ \sum\nolimits_{t=1}^T \log p(\x_t \mid \z_t) \right]\\ &  + \E \left[ \log p(\g_1) + \sum\nolimits_{t=2}^T \log p(\g_t \mid \g_{t-1}, \us_{t-1})\right]\\
    &  - \E \left[\log q(\g_{1:T}, \z_{1:T} \mid \x_{1:T}, \us_{1:T-1})  \right],
    \label{eq:orig_logl}
\end{split}
\end{equation}
where the expectations are taken with respect to samples of $\z_{1:T}$ and $\g_{1:T}$ drawn from $q$.
	
We now consider simplified expressions for the approximate posterior distribution.
The chain rule tells us that:
\begin{equation}
\begin{split}
q& (\g_{1:T}, \z_{1:T} \mid \x_{1:T}, \us_{1:T-1})  = \\
& q(\g_{1:T} \mid \x_{1:T}, \us_{1:T-1})q(\z_{1:T} \mid \g_{1:T}, \x_{1:T}, \us_{1:T-1}).
\end{split}
\end{equation}
As stated previously, given knowledge of $\g_{1:T}$ and $\us_{1:T-1}$, $\z_{1:T}$ are known exactly.
Thus, we know $\log q(\z_{1:T} \mid \g_{1:T}, \x_{1:T}, \us_{1:T-1}) = 0$.
Additionally, we can factorize $q(\g_{1:T} \mid \x_{1:T}, \us_{1:T-1})$ as:
\begin{equation}
\begin{split}
q(\g_{1:T} \mid \x_{1:T}, & \us_{1:T-1}) = q(\g_1 \mid \x_{1:T}, \us_{1:T-1}) \times \\
&  \prod\nolimits_{t=2}^T q(\g_t \mid \g_{1:t-1}, \x_{1:T}, \us_{1:T-1}).
\label{eq:posterior}
\end{split}
\end{equation}
Since all variables $\g_t$ are assumed to be parents of variables $\z_{1:T-1}$, influence can flow from all states and actions to each $\g_t$, and thus the above expression cannot be simplified any further based on conditional independence relationships. 
Taking the logarithm of the quantities in \cref{eq:posterior} and incorporating these into \cref{eq:orig_logl}, we arrive at the following lower bound on the log-likelihood objective:
\begin{equation}
\begin{split}
& \ell \geq  \E_{\z_{1:T} \sim q}  \left[ \sum\nolimits_{t=1}^T \log p(\x_t \mid \z_t) \right] \\
&- \mathcal D_\text{KL} \left[q(\g_1 \mid \x_{1:T}, \us_{1:T-1}) \mid \mid p(\g_1)  \right] - \sum\nolimits_{t=2}^T \mathcal D_\text{KL}^t,
\end{split}
\end{equation}
where $\mathcal D_\text{KL}^t$ represents the KL-divergence between $q(\g_t \mid \g_{1:t-1}, \x_{1:T}, \us_{1:T-1})$ and $p(\g_t \mid \g_{t-1}, \us_{t-1})$ for $t = 2, \ldots, T$.
Thus, we have found a lower bound on our true objective that is comprised of the likelihood of the observed states $\x_{1:T}$ given $\z_{1:T}$, as well as the KL-divergence between the approximate posterior and (conditional) prior distributions over the observations $\g_t$. 
The following section provides a practical training procedure for maximizing this objective.

\subsection{Optimization Procedure}\vspace{-0.25em}

The expectation in the derived lower bound can be estimated through Monte Carlo sampling.
To raise the lower bound, we simultaneously optimize the parameters of six neural networks, which together comprise the Deep Variational Koopman model. The size of each neural network, which is held constant across all experiments, is listed after the name of each model (e.g. $[64, 64]$ would represent a two-layer neural network with 64 neurons in each layer).
\begin{enumerate}
	\item The \emph{Decoder Network} $[64, 32]$ is parameterized by $\bm{\theta}$ and outputs $\bm{\mu}_t$, the mean of a Gaussian distribution over state $\x_t$ given $\z_t$, represented by $p_{\bm{\theta}} (\x_t \mid \z_t)$. We assume that the distribution over $\x_t$ has constant covariance. Hence, maximizing the log-likelihood is equivalent to minimizing the square error between $\bm{\mu}_t$ and $\x_t$.
	\item The \emph{Temporal Encoder Network} $[64]$ is a bidirectional LSTM that maps a sequence of states $\x_{1:T}$ and actions $\us_{1:T-1}$ to a low-dimensional encoding that summarizes the system time evolution.
	\item The \emph{Initial Observation Inference Network} $[64]$ is parameterized by $\bm{\varphi}$ and outputs the parameters to a Gaussian distribution over observation $\g_1$ given the output of the temporal encoder, represented by $q_{\bm{\varphi}} (\g_1 \mid \x_{1:T}, \us_{1:T-1})$.
	\item The \emph{Observation Encoder Network} $[64]$ is a recurrent neural network that takes in observations $\g_{1:t-1}$ and outputs an encoding describing their time evolution. The encoding is updated as more observations are sampled.
	\item The \emph{Observation Inference Network} $[64, 64]$ is parameterized by $\bm{\phi}$ and outputs the parameters to a Gaussian distribution over observation $\g_t$ given the output of the \emph{Temporal} and \emph{Observation Encoder Networks}, represented by $q_{\bm{\phi}} (\g_t \mid \g_{1:t-1}, \x_{1:T}, \us_{1:T-1})$.
	\item The \emph{Conditional Prior Network} $[64, 32]$ is parameterized by $\bm{\psi}$ and outputs the parameters to a Gaussian conditional prior distribution over observation $\g_t$. The output distribution is conditioned on the previous observation and action, and is represented by $p_{\bm{\psi}}(\g_t \mid \g_{t-1}, \us_{t-1})$.  
\end{enumerate}

Given a sequence of states $\x_{1:T}$ and actions $\us_{1:T-1}$, we can sample observations $\g_{1:T}$ from the \textit{Observation Inference Network} and subsequently find the $A$ and $B$ matrices that govern the observation dynamics through $[A \; B] = YZ^\dagger$ and $A^{-1} = (Y - B \Gamma) X^\dagger$, where $\Gamma = [\us_1, \ldots, \us_{T-1}]$ and $X = [\g(\x_1), \ldots, \g(\x_{T-1})]$. Finally, $\z_{1:T}$ can be found through \cref{eq:zs}, and the \textit{Decoder Network} can output state predictions. 

Each time a new set of observations $\g_{1:T}$ is sampled, we obtain a new, globally linear dynamics model.
By sampling many times, we obtain an ensemble of linear models that can provide a distribution over future outcomes for a given system.
This notion of uncertainty can be appealing for a variety of tasks, including prediction and control in circumstances where data is limited.
The next section details how the DVK model can enable uncertainty-aware control.\vspace{-0.5em}

\section{Control}\label{sec:control}

Our goal is to select a sequence of control inputs $\us_{1:H}$ that minimizes $C = \sum_{t=1}^H c(\x_t, \us_t)$, the total incurred cost, where $c(\x_t, \us_t)$ is the instantaneous cost.
Designing controllers for nonlinear systems can be challenging, while many techniques exist for controlling linear systems.
DVK models provide linear dynamics models, which makes them amenable for incorporation into many control frameworks.
Below we outline considerations relating to using DVK models for control.

\subsection{Cost Function}

While the DVK model provides us with linear dynamics, the dynamics are linear in the latent variables $\z_t$.
Thus, we require a cost that is a function of $\z_t$, not $\x_t$.
Specifying such a cost function can be difficult; a common choice is to define the cost to be the $L_2$-distance between the latent representation of the current state and the representation for a goal state.
Such a choice is restrictive but can be justified when the states are represented as visual inputs~\cite{nair2018visual}.

However, it is often easier to specify the cost as a function of the state directly.
For this reason, we define the cost as a function of the state and construct local quadratic approximations of the cost $\hat c(\z_t, \us_t)$ in the latent space.
Building these local approximations requires finding the gradient and Hessian of the state cost with respect to $\z_t$, which in turn requires nonzero second derivatives of the activation functions in the \emph{Decoder Network}, precluding piecewise linear ReLU activations.

\subsection{Optimizing Action Sequences}

We use the Differential Dynamic Programming (DDP) trajectory optimization algorithm to find an action sequence that minimizes the total predicted cost $C$.
The DDP algorithm starts with an action sequence to form an initial reference trajectory, and uses the dynamic programming principle to iteratively update the action sequence and reference trajectory to minimize the predicted cost~\cite{tassa2012synthesis}.
DDP requires locally quadratic approximations to the system dynamics and cost at all points along the reference trajectory.
The DVK model provides us with linear dynamics models and the ability to find locally quadratic approximations to the cost, and thus can be readily incorporated into the DDP algorithm.

A further consideration is that some systems may have constraints on the control inputs.
To account for the presence of such constraints, we optimize the action sequence $\tilde{\us}_{1:H}$, where we define $\us_t = \us_{\max} \text{tanh}(\tilde{\us}_t)$, and $\us_{\max}$ represents the control saturation limits.
Because of the presence of the hyperbolic tangent in this expression, the DDP algorithm requires us to make a quadratic approximation of the system dynamics with respect to the control inputs $\tilde{\us}_{1:H}$.

\subsection{Accounting for Uncertainty}

The standard DDP algorithm assumes the existence of a single, (locally) linear dynamics model.
However, DVK models give us the ability to sample many possible dynamics models, which taken together can encode uncertainty about how a system will evolve in time.
Accounting for such uncertainty can enable more effective control.
Below are two methods we considered for performing uncertainty-aware control.

\paragraph{Optimize for expected cost.} Given $k$ models with $\{[A, B]_i\}_{i=1:k}$ and $\{\z_{1, i}\}_{i=1:k}$, construct an augmented state $\z_{t, \text{aug}}$ that represents the concatenation of the $\z_t$-values across all models. The dynamics of the augmented state will be described by creating a block-diagonal matrix out of the $A$-matrices and stacking the $B$-matrices into a single matrix. A similar procedure can be performed with the cost gradients and Hessians to find quadratic approximations to the cost function along the reference trajectory. The action sequence can be optimized according to the expected cost across all models.

\paragraph{Optimize for worst-case cost.} Given an initial action sequence and $k$ models, find the model predicting the largest cost and use that model to update the state and action trajectory. Subsequently find the model predicting that largest cost under the new action sequence and repeat until convergence.

\subsection{Model Predictive Control}

In the presence of disturbances or model errors, executing an entire action sequence determined through DDP may be inadvisable.
Because DVK dynamics models will not provide perfect predictions for the time evolution of a system, we perform model predictive control (MPC) for closed-loop trajectory planning.
At each time step, we feed the last $T$ observed states and actions into the DVK model to find an ensemble of dynamics models $\{[A, B]_i\}_{i=1:k}$ and initial states $\{\z_{1, i}\}_{i=1:k}$.
Next, we use the DDP procedures outlined above to solve for action sequence $\us_{1:H}$, execute the first action in the sequence, and replan at the next time step.
\vspace{-0.5em}

\section{Experiments}\vspace{-0.25em}

This section evaluates the performance of the Deep Variational Koopman models on benchmark problems for dynamics modeling and control.
We have limited these experiments to low-dimensional problems because it is easier to visualize whether the models have provided reasonable uncertainty estimates.
However, there is no reason why DVK models could not be applied to high-dimensional systems.
Future work will focus on higher-dimensional problems such as fluid flow control, as it has already been shown that Koopman-based approaches can be effective for such applications~\cite{morton2018deep}.

\subsection{Dynamics Modeling}

We evaluate the ability of DVK models to learn dynamics on three benchmark problems: inverted pendulum, cartpole, and acrobot (or double pendulum). 
We use the OpenAI Gym~\cite{brockman2016openai} implementation of each environment, and modify the cartpole and acrobot environments to give them continuous action spaces. 

\subsubsection{Baseline Models}
We benchmark the performance of DVK models against three baseline models.
For a fair comparison, whenever a baseline shares a component with the DVK model, such as the bidirectional LSTM and decoder in the DVBF and LSTM models, we use the exact same hyperparameters across all models.
In our experiments, all latent states were set to be four-dimensional.

The \emph{Deep Variational Bayes Filter} (DVBF) model~\cite{karl2017deep} assumes the presence of latent states $\z_t$ with locally linear dynamics $\z_{t+1} = A_t \z_t + B_t \us_t + C_t \mathbf{w}_t$, where $A_t$, $B_t$, and $C_t$ are functions of the current latent state and $\mathbf{w}_t$ is a noise vector. 
The distribution over $\mathbf{w}_1$ is output by a bidirectional LSTM that encodes information about the sequence of states $\x_{1:T}$, and $\z_1$ is assumed to be a function of $\mathbf{w}_1$.

The \emph{Long Short Term Memory} (LSTM) model propagates a latent state $\z_t$ with a recurrent neural network and uses a decoder network to map $\z_t$ to $\x_t$. 
The latent state at each time step is a function of the previous latent state, the previous control input, and the network hidden state. 
The initial latent state $\z_1$ is drawn from a distribution output by a bidirectional LSTM that encodes information about states $\x_{1:T}$.

We train an ensemble of \num{10} \emph{Multilayer Perceptrons} (MLPs).
Each model is a fully-connected network trained to map the current state and action to the next state.
We can use the range of predictions in the model ensemble as a measure of uncertainty.
Recent work has shown that ensembles of MLPs can serve as probabilistic dynamics models that enable effective control on a variety of tasks~\cite{chua2018deep}.\vspace{-0.25em}

\begin{figure}[t]
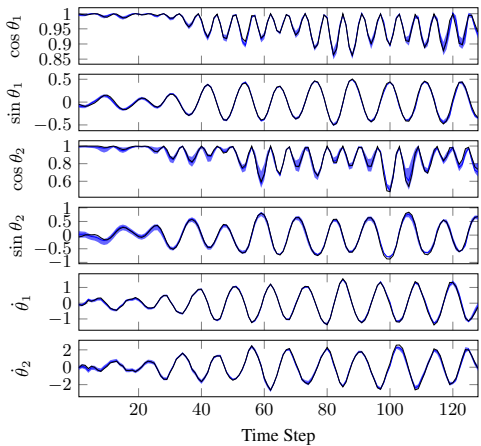

    \begin{center}
        \includestandalone[width=0.75\linewidth]{predictions_kl1}\vspace{-0.5em}
        \caption{Results from the acrobot environment. The first \num{64} steps are reconstruction and the final \num{64} steps are prediction. Black lines indicate true state values, blue lines represent mean predictions, and shaded regions represent the range of predictions across all models.}\vspace{-1.5em}
        \label{fig:acrobot}
    \end{center}
\end{figure}

\begin{figure*}[t]
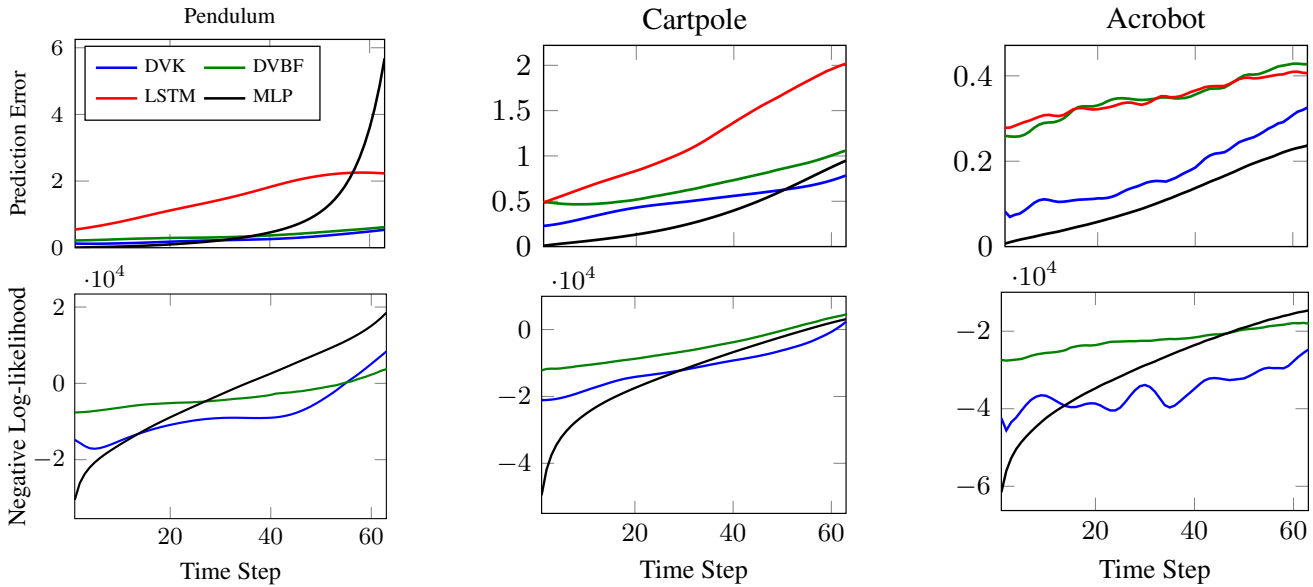


    \begin{subfigure}[t]{0.31 \linewidth}
    \centering
    \includestandalone[width=0.95\linewidth]{pred_pendulum}
    \label{fig:pred_pendulum}
    \end{subfigure}\hfill
    \begin{subfigure}[t]{0.295 \linewidth}
    \centering
    \includestandalone[width=0.95\linewidth]{pred_cartpole}
    \label{fig:pred_cartpole}
    \end{subfigure}\hfill
    \begin{subfigure}[t]{0.295 \linewidth}
    \centering
    \includestandalone[width=0.95\linewidth]{pred_acrobot}
    \label{fig:pred_acrobot}
    \end{subfigure}\\\vspace{-1.5em}

    \begin{subfigure}[t]{0.315 \linewidth}
    \centering
    \includestandalone[width=0.95\linewidth]{likelihood_pendulum}
    \label{fig:likelihood_pendulum}
    \end{subfigure}\hfill
    \begin{subfigure}[t]{0.295 \linewidth}
    \centering
    \hspace*{-0.15em}\includestandalone[width=0.95\linewidth]{likelihood_cartpole}\hspace*{0.15em}
    \label{fig:likelihood_cartpole}
    \end{subfigure}\hfill
    \begin{subfigure}[t]{0.29 \linewidth}
    \centering
    \hspace*{-0.2em}\includestandalone[width=0.98\linewidth]{likelihood_acrobot}\hspace*{0.2em}
    \label{fig:likelihood_acrobot}
    \end{subfigure}\hspace{-0.3em}\vspace{-0.5em}
  \caption{Top: MSE as a function of prediction horizon. Bottom: NLL assigned to the test data (lower is better).}\vspace{-0.5em}
  \label{fig:results}
\end{figure*}

\begin{table*}[h]
 \centering
 \begin{tabular}{ccccccc}
  \toprule
    \multirow{2}{*}{Number of Models} & \multicolumn{2}{c}{Cost} & \multicolumn{2}{c}{Fraction of Time Vertical} & \multicolumn{2}{c}{Falls per Trial} \\
    \cmidrule{2-7}
    & Worst Case & Expected & Worst Case & Expected & Worst Case & Expected\\
    \midrule
    $1$ & - & $370.7$ & - & $0.632$ & - & $0.353$ \\
    $5$ & $307.5$ & $285.7$ & $0.710$ & $0.732$ & $0.312$ & $0.111$ \\
    $10$ & $291.9$ & $274.3$ & $0.733$ & $0.748$ & $0.218$ & $0.063$ \\
    $30$ & $\mathbf{251.6}$ & $270.9$ & $\mathbf{0.771}$ & $0.748$ & $0.061$ & $\mathbf{0.055}$ \\
    \bottomrule
 \end{tabular}\vspace{-0.5em}
 \caption{Control Performance}\vspace{-1em}
 \label{table:control}
\end{table*}

\subsubsection{Training Details}

All models were implemented in TensorFlow~\cite{abadi2015tensorflow}.
In each environment, \num{1000} trials were run for \num{256} time steps with random control inputs.
Models were trained on subsequences of states and actions extracted from the trial data.
Knowledge about a sequence of states $\x_{1:T}$ is required to sample the initial latent state in the DVK, DVBF, and LSTM models (and the dynamics model for DVK).
For this reason, we must draw a distinction between \emph{reconstruction}, in which a model simulates the evolution of states $\x_{1:T}$ about which it already has knowledge, and \emph{prediction}, in which a model predicts the evolution of states $\x_{T+1:T+H}$.

In calculating $Z^\dagger$ and $X^\dagger$, a small scalar value may need to be added to the diagonal entries of $Z$ and $X$ to avoid conditioning issues, meaning that the $A^{-1}$ matrix  found by the DVK model might not be the true inverse of $A$.
If the model is trained only for reconstruction, in which it simulates the system backward in time with $A^{-1}$, it may not perform well in prediction when it uses $A$ for the forward-time dynamics.
To ensure that the system can be simulated accurately with both $A$ and $A^{-1}$, we train the DVK model to minimize the sum of the reconstruction \emph{and} prediction errors for states $\x_{1:T+H}$, where we often set $T = H$.
We found that employing a similar training procedure inhibited learning for the LSTM model, but did improve the predictive performance of the DVBF, and as such all DVBF results are from models trained in this manner.

\subsubsection{Results}

We evaluate the trained models on \num{5000} \num{64}-step test sequences from each environment.
For each test sequence, we generate \num{10} predictions with each model.
For each prediction, the DVBF and LSTM models sample different initial latent states $\z_1$, and the DVK model samples different values of $\z_T$ and dynamics matrices.
We obtain distinct predictions from the MLP ensemble by generating recursive predictions with each trained model.
\Cref{fig:acrobot} provides a qualitative picture of the DVK model's ability to simulate the dynamics on one test sequence.
We can observe strong agreement between the model's predictions and the true time evolution of the system, with higher uncertainty present near local minima and maxima.

The predictive performance of the models are quantified according to two metrics: (1) mean squared error (MSE) as a function of prediction horizon, averaged across the \num{10} predictions and \num{5000} trials, and (2) negative log-likelihood (NLL) of the test data as a function of prediction horizon, summed across trials. 
The likelihood is calculated by fitting a Gaussian distribution to the \num{10} predictions generated by each model and determining the probability density that distribution assigns to the true state value.
\Cref{fig:results} shows model performance according to these metrics on the three studied environments.
The likelihood results for the LSTM model are omitted because its fitted distributions assigned zero likelihood to some of the test data, which corresponds to infinite negative log-likelihood.

The MLP ensemble performs quite well, achieving low prediction error and assigning high likelihood to the test data.
However, the results for the pendulum problem, where the prediction error grows exponentially, illustrate one drawback of using models trained to make single-step predictions.
The prediction errors for such models can grow exponentially when they are used to make multi-step predictions due to errors compounding over time~\cite{venkatraman2015improving}.
In fact, over a horizon of \num{128} time steps in the pendulum environment the mean-squared prediction error for the MLP ensemble grows to values on the order of $10^6$.
The DVK model achieves competitive performance with the MLP ensemble, while not suffering from the same instabilities and also providing linear dynamics models that can be used more easily for control.

The DVBF outperforms the LSTM baseline, and attains performance that is often close to that of the DVK model, but at a much higher computational cost.
The DVK model computes a single dynamics model that it uses to propagate the latent state for all time steps, while the DVBF must compute a new dynamics model, which is a nonlinear function of the current latent state, at each time step.
Therefore, the computational graph for the DVBF takes significantly longer to compile, and furthermore in our experiments the time required to perform a forward and backward pass during training was approximately an order or magnitude longer for the DVBF.\vspace{-0.5em}

\subsection{Control}\vspace{-0.25em}

We evaluate the effectiveness of the control procedure detailed in \cref{sec:control} on the inverted pendulum environment, in which the goal is to swing up and balance an underactuated pendulum.
The cost function penalizes deviations of the pendulum from vertical, as well as nonzero angular velocities and control inputs.
In each episode, the pendulum is initialized in a random state and the system is simulated for \num{256} time steps.

We ran \num{50} trials with random control inputs and trained a DVK model on the trial data with a reconstruction and prediction horizon of $T=H=16$.
Because the original dataset did not contain many instances where the pendulum was near the goal state, we ran \num{20} additional trials where the actions were selected through MPC optimizing for expected cost with five sampled models.
We then finetuned the DVK model on data from these trials before carrying out the final experiments.

The results for different ensemble sizes and optimization procedures, taken from \num{1000} seeded trials, can be found in \cref{table:control}.
We quantify performance according to (1) the average cost incurred in each trial, (2) the fraction of the time the pendulum was vertical ($\theta \in [-\pi/8, \pi/8]$) across all trials, and (3) the average number of falls per trial.
A fall is classified as a scenario where $\theta \not\in [-\pi/8, \pi/8]$ after being in the interval for more than 20 time steps. 
The best performance according to each metric is highlighted in bold.

The results show a clear benefit from sampling more models.
The trend in performance improvement is more pronounced for the worst-case optimization scheme; when optimizing for expected cost we do see a benefit to sampling more models, but with diminishing returns.
The best performance is obtained with \num{30} models, with the lowest average cost achieved through a worst-case optimization procedure.
However, optimizing for expected cost leads to much better performance for smaller model ensembles, and thus could be a preferable approach if the goal is to obtain satisfactory performance while keeping the number of sampled models low.\vspace{-0.25em}

\section{Conclusions} \vspace{-0.25em}

We introduced the Deep Variational Koopman model, a method for inferring Koopman observations and sampling ensembles of linear dynamics models that can be used for prediction and control.
We demonstrated that DVK models were able to perform accurate, long-term prediction on a series of benchmark tasks, and that accounting for the uncertainty encoded by multiple sampled models improved controller performance on the inverted pendulum task.
Future work will focus on applying the DVK models to higher-dimensional problems and more complex tasks, such as fluid flow control.
Source code associated with this project can be found at {\tt\small https://github.com/sisl/variational\char`_koopman}.\vspace{-0.25em}

\section*{Acknowledgments}\vspace{-0.25em}
This material is based upon work supported by the National Science Foundation Graduate Research Fellowship Program under Grant No. DGE- 114747 and the Stanford Energy Alliance.
The authors would like to thank Zac Manchester for valuable feedback.

\bibliographystyle{named}
\bibliography{ref}

\end{document}

%% file: graph.tex
\begin{tikzpicture}
    \tikzset{
      value/.style={align=center, minimum height=8mm, minimum width=8mm, font={\small}, circle, draw=black},
      arrow/.style={solid, thick, ->,>=stealth',shorten >=1pt},
    }
    \node[value] (gt) {$\mathbf{g}_t$};
    \node[value] (gtp1) [right=1.0 of gt] {$\mathbf{g}_{t+1}$};
    \node[value] (ut) [below=1.0 of gt] {$\mathbf{u}_t$};
    \draw[arrow] (gt) to (gtp1); 
    \draw[arrow] (ut) to (gtp1);

    \node[value] (zt) [right=6.0 of gt] {$\mathbf{z}_{t}$};
    \node[value] (ztm1) [right=1.0 of zt] {$\mathbf{z}_{t-1}$};
    \node[value] (xt) [below=1.5 of zt] {$\mathbf{x}_t$};
    \draw[arrow] (ztm1) to (zt);
    \draw[arrow] (zt) to (xt);
    \node[value] (plate_anchor)[above=1.5 of zt, draw=none]{};
    \node[value] (gtnode) [left=1.0 of plate_anchor] {$\mathbf{g}_t$};
    \node[value] (utnode) [right=1.0 of plate_anchor] {$\mathbf{u}_t$};
    \node[value] (gtphantom)[right=-0.65 of gtnode, draw=none]{};
    \node[value] (utphantom)[right=-0.3 of utnode, draw=none]{};
    \plate  {plateg} {(gtphantom)} {$t = 1:T$};
    \plate  {plateu} {(utphantom)} {$t = 1:T-1$};
    \draw[arrow] (plateg) to (zt);
    \draw[arrow] (plateu) to (zt);

\end{tikzpicture}

%% file: predictions_kl1.tex
\begin{tikzpicture}[font=\footnotesize]
\begin{groupplot}[group style={group size=1 by 6, vertical sep=0.5em},height=\linewidth,width=\linewidth]
\nextgroupplot[ 
	ylabel=$\cos\theta_1$,
    ylabel shift = 2pt,
	height=0.3\linewidth,
	width=\linewidth,
	xmin=1, xmax=128, xticklabels={,,}]

\addplot +[name path=pluserror,draw=none,no markers,forget plot]
    table[x=t,y=max1, col sep=comma]{prediction_ranges_high_kl.csv};

\addplot [name path=minuserror,draw=none,no markers,forget plot]
    table [x=t,y=min1, col sep=comma] {prediction_ranges_high_kl.csv};

\addplot [forget plot,fill=blue,opacity=0.6]
    fill between[on layer={},of=pluserror and minuserror];

\addplot +[blue,thin,no markers]
    table[x=t,y=mean1, col sep=comma] {prediction_ranges_high_kl.csv};

\addplot +[black,thin,no markers]
    table[x=t,y=true1, col sep=comma] {prediction_ranges_high_kl.csv};

\nextgroupplot[
    ylabel=$\sin\theta_1$,
    ylabel shift = -1pt,
    height=0.3\linewidth,
    width=\linewidth,
    xmin=1, xmax=128, xticklabels={,,}]

\addplot +[name path=pluserror,draw=none,no markers,forget plot]
    table[x=t,y=max2, col sep=comma]{prediction_ranges_high_kl.csv};

\addplot [name path=minuserror,draw=none,no markers,forget plot]
    table [x=t,y=min2, col sep=comma] {prediction_ranges_high_kl.csv};

\addplot [forget plot,fill=blue,opacity=0.6]
    fill between[on layer={},of=pluserror and minuserror];

\addplot +[blue,thin,no markers]
    table[x=t,y=mean2, col sep=comma] {prediction_ranges_high_kl.csv};

\addplot +[black,thin,no markers]
    table[x=t,y=true2, col sep=comma] {prediction_ranges_high_kl.csv};

\nextgroupplot[ 
    ylabel=$\cos\theta_2$,
    ylabel shift = 5pt,
    height=0.3\linewidth,
    width=\linewidth,
    xmin=1, xmax=128, xticklabels={,,}]

\addplot +[name path=pluserror,draw=none,no markers,forget plot]
    table[x=t,y=max3, col sep=comma]{prediction_ranges_high_kl.csv};

\addplot [name path=minuserror,draw=none,no markers,forget plot]
    table [x=t,y=min3, col sep=comma] {prediction_ranges_high_kl.csv};

\addplot [forget plot,fill=blue,opacity=0.6]
    fill between[on layer={},of=pluserror and minuserror];

\addplot +[blue,thin,no markers]
    table[x=t,y=mean3, col sep=comma] {prediction_ranges_high_kl.csv};

\addplot +[black,thin,no markers]
    table[x=t,y=true3, col sep=comma] {prediction_ranges_high_kl.csv};

\nextgroupplot[
    ylabel=$\sin\theta_2$,
    ylabel shift = -2pt,
    height=0.3\linewidth,
    width=\linewidth,
    xmin=1, xmax=128, xticklabels={,,}]

\addplot +[name path=pluserror,draw=none,no markers,forget plot]
    table[x=t,y=max4, col sep=comma]{prediction_ranges_high_kl.csv};

\addplot [name path=minuserror,draw=none,no markers,forget plot]
    table [x=t,y=min4, col sep=comma] {prediction_ranges_high_kl.csv};

\addplot [forget plot,fill=blue,opacity=0.6]
    fill between[on layer={},of=pluserror and minuserror];

\addplot +[blue,thin,no markers]
    table[x=t,y=mean4, col sep=comma] {prediction_ranges_high_kl.csv};

\addplot +[black,thin,no markers]
    table[x=t,y=true4, col sep=comma] {prediction_ranges_high_kl.csv};

\nextgroupplot[ 
    ylabel=$\dot\theta_1$,
    ylabel shift = 2pt,
    height=0.3\linewidth,
    width=\linewidth,
    xmin=1, xmax=128, xticklabels={,,}]

\addplot +[name path=pluserror,draw=none,no markers,forget plot]
    table[x=t,y=max5, col sep=comma]{prediction_ranges_high_kl.csv};

\addplot [name path=minuserror,draw=none,no markers,forget plot]
    table [x=t,y=min5, col sep=comma] {prediction_ranges_high_kl.csv};

\addplot [forget plot,fill=blue,opacity=0.6]
    fill between[on layer={},of=pluserror and minuserror];

\addplot +[blue,thin,no markers]
    table[x=t,y=mean5, col sep=comma] {prediction_ranges_high_kl.csv};

\addplot +[black,thin,no markers]
    table[x=t,y=true5, col sep=comma] {prediction_ranges_high_kl.csv};

\nextgroupplot[
    xlabel = Time Step,
    ylabel=$\dot\theta_2$,
    ylabel shift = 2pt,
    height=0.3\linewidth,
    width=\linewidth,
    xmin=1, xmax=128]

\addplot +[name path=pluserror,draw=none,no markers,forget plot]
    table[x=t,y=max6, col sep=comma]{prediction_ranges_high_kl.csv};

\addplot [name path=minuserror,draw=none,no markers,forget plot]
    table [x=t,y=min6, col sep=comma] {prediction_ranges_high_kl.csv};

\addplot [forget plot,fill=blue,opacity=0.6]
    fill between[on layer={},of=pluserror and minuserror];

\addplot +[blue,thin,no markers]
    table[x=t,y=mean6, col sep=comma] {prediction_ranges_high_kl.csv};

\addplot +[black,thin,no markers]
    table[x=t,y=true6, col sep=comma] {prediction_ranges_high_kl.csv};
    
\end{groupplot}
\end{tikzpicture}

%% file: pred_pendulum.tex
\begin{tikzpicture}[font=\scriptsize]
\begin{axis}[
	ylabel=Prediction Error,
	title=Pendulum,
	title style={yshift=-1ex,},
	xticklabels={,,},
	ylabel shift = 1pt,
	height=0.7\linewidth,
	width=0.9\linewidth,
	xmin=1, xmax=63, ymin=0,
	legend pos = north west,
	legend columns = 2,
	legend style={font=\tiny},
	legend image post style={scale=0.75}]

\addplot +[blue,thick,no markers]
    table[x=t,y=error, col sep=comma] {vk_pendulum_pred_error.csv};
    
\addplot +[colorGreenish,thick,no markers]
    table[x=t,y=error, col sep=comma] {bf_pendulum_pred_error.csv};
    
\addplot +[red,thick,no markers]
    table[x=t,y=error, col sep=comma] {rnn_pendulum_pred_error.csv};

\addplot +[black,thick,no markers]
    table[x=t,y=error, col sep=comma] {mlp_pendulum_pred_error.csv};

\legend{DVK, DVBF, LSTM, MLP}
\end{axis}
\end{tikzpicture}

%% file: pred_cartpole.tex
\begin{tikzpicture}[font=\footnotesize]
\begin{axis}[
	height=0.7\linewidth,
	title=Cartpole,
	title style={yshift=-1.5ex,},
	xticklabels={,,},
	width=0.9\linewidth,
	xmin=1, xmax=63, ymin=0,
	legend pos = north west,
	legend columns = 1,
	legend image post style={scale=0.75}]

\addplot +[blue,thick,no markers]
    table[x=t,y=error, col sep=comma] {vk_cartpole_pred_error.csv};
    
\addplot +[colorGreenish,thick,no markers]
    table[x=t,y=error, col sep=comma] {bf_cartpole_pred_error.csv};
    
\addplot +[red,thick,no markers]
    table[x=t,y=error, col sep=comma] {rnn_cartpole_pred_error.csv};

\addplot +[black,thick,no markers]
    table[x=t,y=error, col sep=comma] {mlp_cartpole_pred_error.csv};

\end{axis}
\end{tikzpicture}

%% file: pred_acrobot.tex
\begin{tikzpicture}[font=\footnotesize]
\begin{axis}[
	height=0.7\linewidth,
	title=Acrobot,
	title style={yshift=-1ex,},
	xticklabels={,,},
	width=0.9\linewidth,
	xmin=1, xmax=63, ymin=0,
	legend pos = north west,
	legend columns = 1,
	legend image post style={scale=0.75}]

\addplot +[blue,thick,no markers]
    table[x=t,y=error, col sep=comma] {vk_acrobot_pred_error.csv};
    
\addplot +[colorGreenish,thick,no markers]
    table[x=t,y=error, col sep=comma] {bf_acrobot_pred_error.csv};
    
\addplot +[red,thick,no markers]
    table[x=t,y=error, col sep=comma] {rnn_acrobot_pred_error.csv};

\addplot +[black,thick,no markers]
    table[x=t,y=error, col sep=comma] {mlp_acrobot_pred_error.csv};

\end{axis}
\end{tikzpicture}

%% file: likelihood_pendulum.tex
\begin{tikzpicture}[font=\footnotesize]
\begin{axis}[
	xlabel=Time Step,
	ylabel=Negative Log-likelihood,
	ylabel shift = -6pt,
	height=0.8\linewidth,
	width=\linewidth,
	xmin=1, xmax=63,
	legend pos = north west,
	legend columns = 1,
	legend image post style={scale=0.75}]

\addplot +[blue,thick,no markers]
    table[x=t,y=likelihood, col sep=comma] {vk_pendulum_likelihood.csv};
    
\addplot +[colorGreenish,thick,no markers]
    table[x=t,y=likelihood, col sep=comma] {bf_pendulum_likelihood.csv};

\addplot +[black,thick,no markers]
    table[x=t,y=likelihood, col sep=comma] {mlp_pendulum_likelihood.csv};

\end{axis}
\end{tikzpicture}

%% file: likelihood_cartpole.tex
\begin{tikzpicture}[font=\footnotesize]
\begin{axis}[
	xlabel=Time Step,
	height=0.8\linewidth,
	width=\linewidth,
	xmin=1, xmax=63,
	legend pos = north west,
	legend columns = 1,
	legend image post style={scale=0.75}]

\addplot +[blue,thick,no markers]
    table[x=t,y=likelihood, col sep=comma] {vk_cartpole_likelihood.csv};
    
\addplot +[colorGreenish,thick,no markers]
    table[x=t,y=likelihood, col sep=comma] {bf_cartpole_likelihood.csv};

\addplot +[black,thick,no markers]
    table[x=t,y=likelihood, col sep=comma] {mlp_cartpole_likelihood.csv};

\end{axis}
\end{tikzpicture}

%% file: likelihood_acrobot.tex
\begin{tikzpicture}[font=\footnotesize]
\begin{axis}[
	xlabel=Time Step,
	height=0.8\linewidth,
	width=\linewidth,
	xmin=1, xmax=63,
	legend pos = north west,
	legend columns = 1,
	legend image post style={scale=0.75}]

\addplot +[blue,thick,no markers]
    table[x=t,y=likelihood, col sep=comma] {vk_acrobot_likelihood.csv};
    
\addplot +[colorGreenish,thick,no markers]
    table[x=t,y=likelihood, col sep=comma] {bf_acrobot_likelihood.csv};

\addplot +[black,thick,no markers]
    table[x=t,y=likelihood, col sep=comma] {mlp_acrobot_likelihood.csv};

\end{axis}
\end{tikzpicture}